\documentclass[runningheads]{llncs}

\usepackage[english]{babel}      
\usepackage{graphicx}
\usepackage{hyperref}
\usepackage{subfig}
\usepackage{times}

\graphicspath{{figures/}}

\begin{document}

\title{Visualizing Convolutional Neural Networks \\ to Improve Decision Support \\ for Skin Lesion Classification}

\titlerunning{Visualizing Convolutional Neural Networks}

\author{Pieter Van Molle \and
	Miguel De Strooper \and
	Tim Verbelen \and \\
	Bert Vankeirsbilck \and
    Pieter Simoens \and
    Bart Dhoedt
}
\authorrunning{P. Van Molle et al.}
\institute{IDLab, Department of Information Technology at Ghent University - imec
\email{firstname.lastname@ugent.be}}
\maketitle 
\begin{abstract}
Because of their state-of-the-art performance in computer vision, CNNs are becoming increasingly popular in a variety of fields, including medicine. However, as neural networks are black box function approximators, it is difficult, if not impossible, for a medical expert to reason about their output. This could potentially result in the expert distrusting the network when he or she does not agree with its output. In such a case, explaining why the CNN makes a certain decision becomes valuable information. In this paper, we try to open the black box of the CNN by inspecting and visualizing the learned feature maps, in the field of dermatology. We show that, to some extent, CNNs focus on features similar to those used by dermatologists to make a diagnosis. However, more research is required for fully explaining their output.

\end{abstract}

\section{Introduction}
Over the past few years, deep neural network architectures---convolutional architectures in particular---have time and again beaten state-of-the-art on large-scale image recognition tasks \cite{krizhevsky2012imagenet,simonyan2014very,szegedy2016rethinking,he2016deep}. As a result, the application of convolutional neural networks (CNN) has become increasingly popular in a variety of fields. In medicine, deep learning is used as a tool to assist professionals of various subfields in their diagnoses, such as histopathology \cite{litjens2016deep}, oncology \cite{cirecsan2013mitosis,fakoor2013using,wang2016deep}, pulmonology \cite{hua2015computer,sun2016computer}, etc\footnote{We refer the reader to \cite{litjens2017survey} for an in-depth survey on deep learning in medical analysis.}. In the subfield of dermatology, CNNs have been applied to the problem of skin lesion classification, based on dermoscopy images, where they set a new state-of-the-art benchmark, matching---or even surpassing---medical expert performance \cite{codella2017deep,esteva2017dermatologist,Haenssle18}.

The challenge remains, however, to understand the reasoning behind the decisions made by these networks, since they are essentially black box function approximators. This poses a problem when a neural network outputs a diagnosis, different from the diagnosis made by the medical expert, as there is no human interpretable reasoning behind the neural networks' diagnosis. In such a case, visualizations of the network could serve as a reasoning tool to the expert.

In this paper, we train a CNN for binary classification on a skin lesion dataset, and inspect the features learned by the network, by visualizing its feature maps. In the next section, we first give an overview of the different visualization strategies for inspecting CNNs. Section 3 describes our CNN architecture and training procedure. In Section 4 we present and discuss the learned CNN features and we conclude the paper in Section 5.

\section{Related Work}

In \cite{zeiler2014visualizing}, the authors propose a visualization technique to give some insight into the function of the intermediate feature maps of a trained CNN, by attaching a deconvolutional network to each of its convolutional layers. While a CNN maps the input from the image space to a feature space, a deconvolutional network does the opposite (mapping from a feature space back to the image space), by reversing its operations. This is done by a series of unpooling, rectifying and filtering operations. The authors use a deconvolutional network to visualize the features that result in the highest activations in a given feature map. Furthermore, they evaluate the sensitivity of a feature map, to the occlusion of a certain part of the input image, and the effect it has on the class score for the correct class.

Two other visualization techniques are presented in \cite{simonyan2013deep} that are based on optimization. The first technique iteratively generates a canonical image representing a class of interest. To generate this image, the authors start from a zero image and pass it through a trained CNN. Optimization is done by means of the back-propagation algorithm, by calculating the derivative of the class score, with respect to the image, while keeping the parameters of the network fixed. The second technique aims to visualize the image-specific class saliency. For a given input image and a class of interest, they calculate the derivative of the class score, with respect to the input image. The per-pixel derivatives of the input image give an estimate of the importance of these pixels regarding the class score. More specifically, the magnitude of the derivate indicates which pixels affect the class score the most when they are changed.

Concluding, typical visualization techniques either generate a single output image, in case of the feature visualization and the generation of the class representative, or function at the pixel level of the input image, in case of the region occlusion and the image-specific class saliency visualization. However, dermatologists typically scan a lesion for the presence of different individual features, such as asymmetry, border, color and structures, i.e. the so-called ABCD-score~\cite{nachbar1994abcd}. Therefore, we inspect and visualize the intermediary feature maps of the CNN on a per-image basis, aiming to provide more familiar insights to dermatologists.

\section{Architecture and Training}

A common approach is to use a CNN pre-trained on a large image database such as ImageNet and then fine-tune this on the target dataset \cite{Haenssle18}. The drawback is that this CNN will also contain a lot of uninformative filters (e.g. for classifying cats and dogs) for the domain at hand. Therefore we chose to train a basic CNN from scratch, but in principle our visualization approach can work for any CNN.

Our CNN consists of 4 convolutional blocks, each formed by 2 convolutional layers followed by a max pooling operation. The convolutional layers in each block have a kernel size of $3 \times 3$, and have respectively 8, 16, 32 and 64 filters. This is followed by 3 fully connected layers with 2056, 1024 and 64 hidden units. All layers have rectified linear units (ReLU) as non-linearity. 

We use data from the publicly available ISIC Archive\footnote{\url{https://isic-archive.com/}}, to compose a training set of 12,838 dermoscopy images, spread over two classes (11,910 benign lesions, 928 malignant lesions). In a preprocessing step, the images are downscaled to a resolution of $300 \times 300$ pixels, and RGB values are normalized between 0 and 1. We augment our training set by taking random crops of $224 \times 224$ pixels, and further augment each crop by rotating (angle sampled uniformly between 0 and $2 \pi$), randomly flipping horizontally and/or vertically, adjusting brightness (factor sampled uniformly between -0.5 and 0.5), contrast (factor sampled uniformly between -0.7 and 0.7), hue (factor sampled uniformly between -0.02 and 0.02) and saturation (factor sampled uniformly between 0.7 and 1.5).

We have trained the network for 192 epochs, with mini-batches of size 96 and used the Adam algorithm \cite{kingma2014adam} to update the parameters of the network, with an initial learning rate of $10^{-4}$ and an exponential decay rate for the first and second order momentum of respectively 0.9 and 0.999. We have evaluated the performance of the resulting CNN on a hold-out test set, comprised of 600 dermoscopy images (483 benign lesions, 117 malignant lesions), achieving an AUC score of 0.75.

\section{Feature Map Visualization}

For each feature map of the CNN, we created a visualization by rescaling the feature map to the input size and overlaying the activations mapped to a transparent green color (darker green = higher activation). We identify each visualization by the convolutional layer number (0..7) and filter number. Next we inspected all visualizations and tried to relate these to typical features dermatologists scan for. Especially the last two convolutional layers of the CNN (6,7) give us some insights into which image regions grasp the attention of the CNN.

\begin{figure}[t!]
	\begin{center}		
		\subfloat[7, 28]{
		\begin{tabular}{cc}
			\includegraphics[width = 0.21\textwidth]{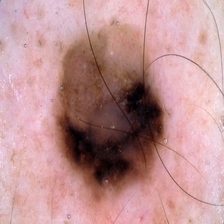} &
			\includegraphics[width = 0.21\textwidth]{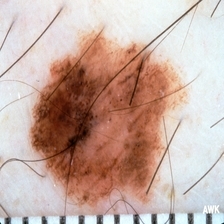} 
			\\
			\includegraphics[width = 0.21\textwidth]{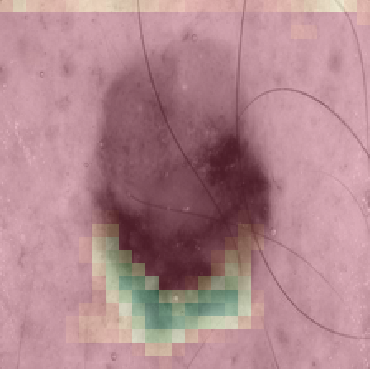} &
			\includegraphics[width = 0.21\textwidth]{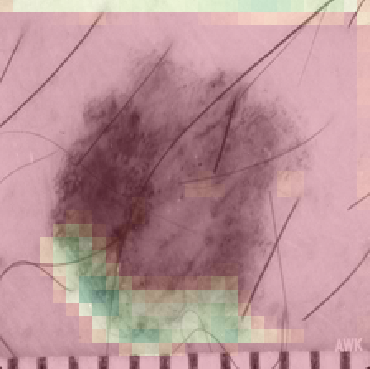} 
		\end{tabular}		
		}
		\subfloat[7, 32]{
		\begin{tabular}{cc}
			\includegraphics[width = 0.21\textwidth]{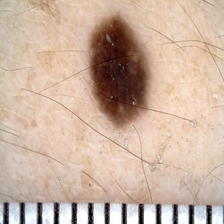} &
			\includegraphics[width = 0.21\textwidth]{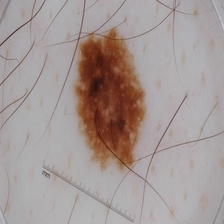}  
			\\
			\includegraphics[width = 0.21\textwidth]{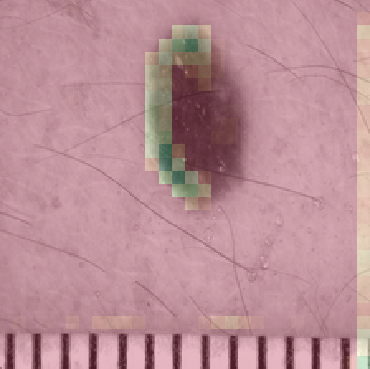} &
			\includegraphics[width = 0.21\textwidth]{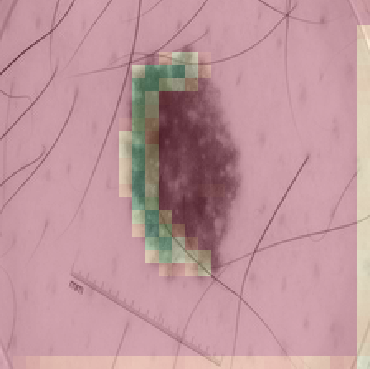} 
		\end{tabular}		
		}
		\caption{Feature maps with high activations on lesion borders, specializing on the border location. For example, filter (a) activates on the bottom border, while filter (b) activates on the left border. \label{fig:borders}}
	\end{center}
\end{figure}

\textbf{Borders} Irregularities in the border of a skin lesion could indicate a malignant lesion. The feature maps shown in Fig.~\ref{fig:borders} both have high activations on the border of a skin lesion, but on different parts of the border. The first one (a) detects the bottom border of a lesion, while the second one (b) detects the left border.

\begin{figure}[t!]
	\begin{center}		
		\subfloat[6, 17]{
		\begin{tabular}{cc}
			\includegraphics[width = 0.21\textwidth]{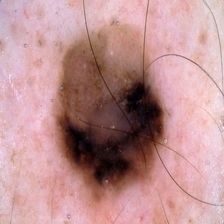} &
			\includegraphics[width = 0.21\textwidth]{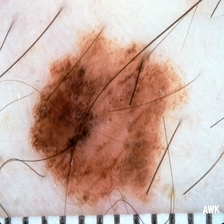} 
			\\
			\includegraphics[width = 0.21\textwidth]{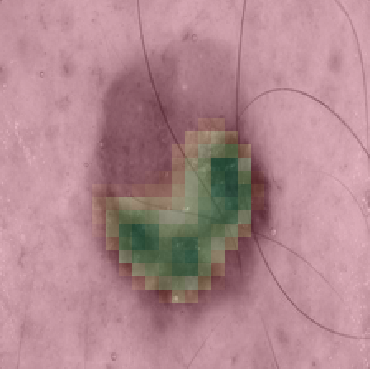} &
			\includegraphics[width = 0.21\textwidth]{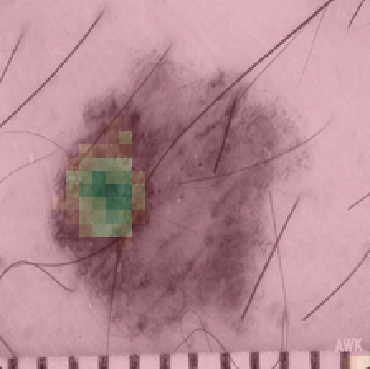} 
		\end{tabular}		
		}	
		\subfloat[6, 58]{
		\begin{tabular}{cc}
			\includegraphics[width = 0.21\textwidth]{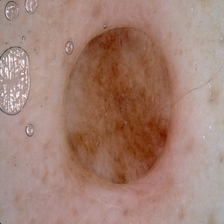} &
			\includegraphics[width = 0.21\textwidth]{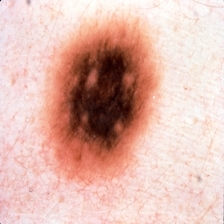}  
			\\
			\includegraphics[width = 0.21\textwidth]{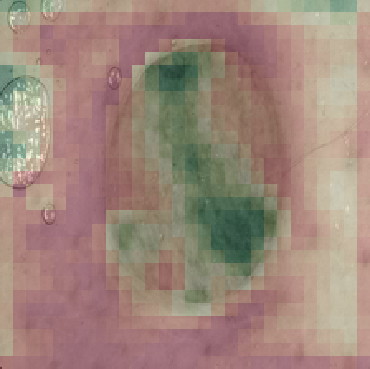} &
			\includegraphics[width = 0.21\textwidth]{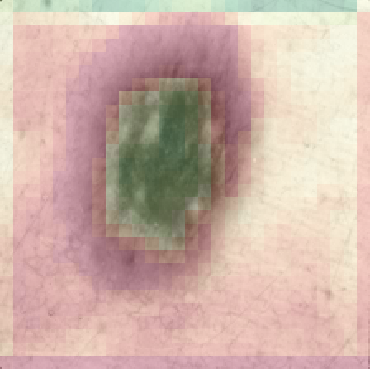} 
		\end{tabular}		
		}
		\caption{Feature maps with high activations on darker regions within the lesion, indicating a non-uniformity in the color of the lesion. \label{fig:colors}}
	\end{center}
\end{figure}

\textbf{Color} The same reasoning tends to apply to the colors inside the lesion. A lesion that has a uniform color is usually benign, while major irregularities in color could be a sign of a malignant lesion. The feature maps shown in Fig.~\ref{fig:colors} have a high activation when a darker region is present in the lesion, implying a non-uniform color.

\begin{figure}[t!]
	\begin{center}
		\subfloat[6, 44]{
		\begin{tabular}{cc}
			\includegraphics[width = 0.21\textwidth]{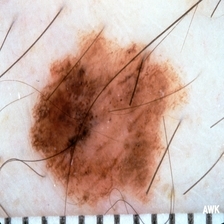} &
			\includegraphics[width = 0.21\textwidth]{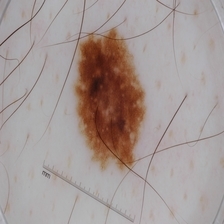}  
			\\
			\includegraphics[width = 0.21\textwidth]{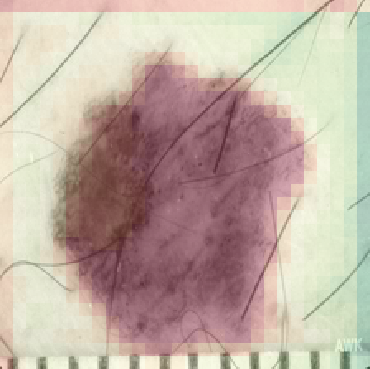} &
			\includegraphics[width = 0.21\textwidth]{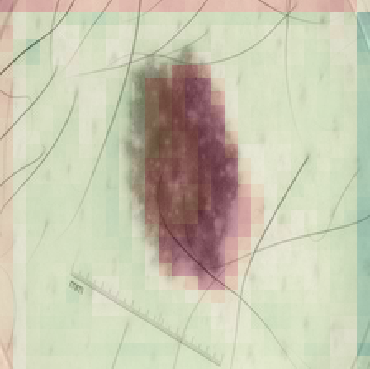} 
		\end{tabular}		
		}		
		\subfloat[7, 33]{
		\begin{tabular}{cc}
			\includegraphics[width = 0.21\textwidth]{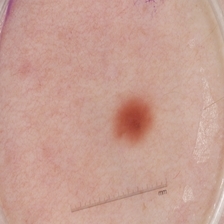} &
			\includegraphics[width = 0.21\textwidth]{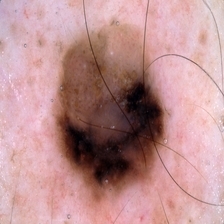} 
			\\
			\includegraphics[width = 0.21\textwidth]{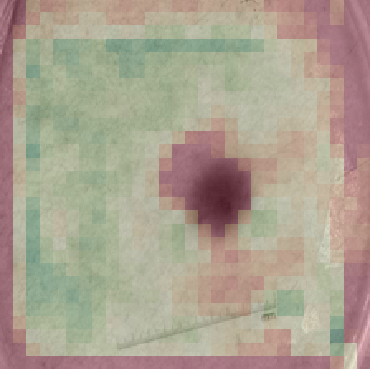} &
			\includegraphics[width = 0.21\textwidth]{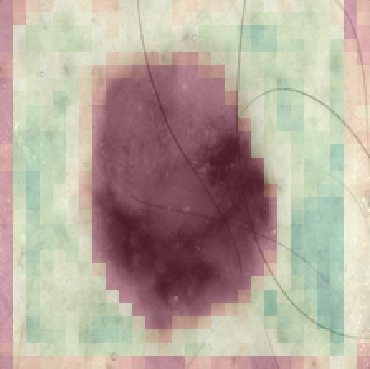} 
		\end{tabular}		
		}	
		\caption{Feature maps with high activations on skin types. For example, filter (a) activates on pale skin, while filter (b) activates on pink skin texture. \label{fig:skin}}
	\end{center}
\end{figure}

\textbf{Skin Type} People with a lighter skin are more prone to sunburns, which can increase the development of malignant lesions on their skin. Therefore, a dermatologist takes a patient's skin type into account when examining his or her lesions. The same goes for the feature maps shown in Fig.~\ref{fig:skin}. The first feature map (a) has high activations on white-pale skin. The second one (b) has high activations on a more pinkish skin with vessel-like structures.

\begin{figure}[t!]
	\begin{center}
		\begin{tabular}{ccc} 
			\includegraphics[width = 0.21\textwidth]{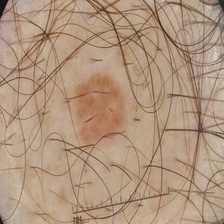} &
			\includegraphics[width = 0.21\textwidth]{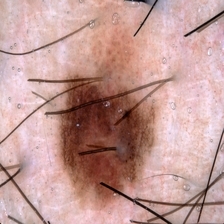} &
			\includegraphics[width = 0.21\textwidth]{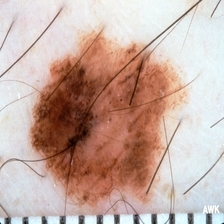} 
			\\
			\includegraphics[width = 0.21\textwidth]{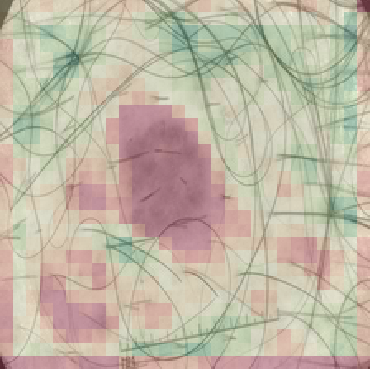} &
			\includegraphics[width = 0.21\textwidth]{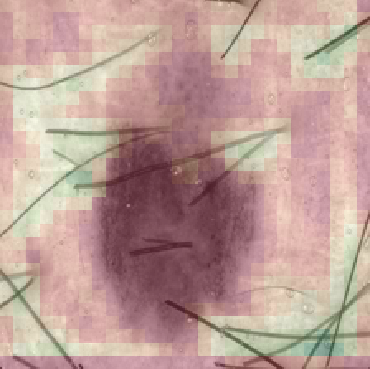} &
			\includegraphics[width = 0.21\textwidth]{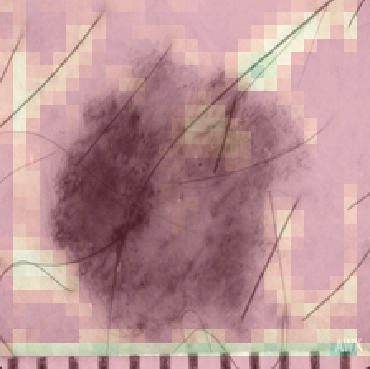} 
		\end{tabular}
		\caption{A feature map (7, 8) with high activations on hair-like structures. \label{fig:hair}}
	\end{center}
\end{figure}

\textbf{Hair} The CNN also learns feature maps that, from a dermatologist viewpoint, have no impact on the diagnosis. For example, the feature map in Fig.~\ref{fig:hair} has high activations on hair-like structures. 

\begin{figure}[t!]
	\begin{center}
		\begin{tabular}{ccc} 
			\includegraphics[width = 0.21\textwidth]{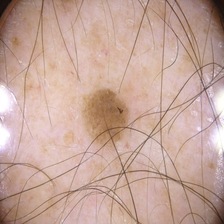} &
			\includegraphics[width = 0.21\textwidth]{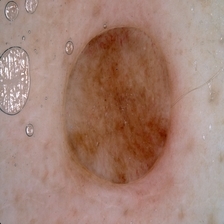} &
			\includegraphics[width = 0.21\textwidth]{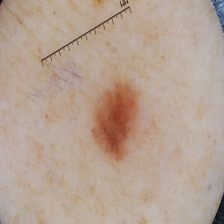} 
			\\
			\includegraphics[width = 0.21\textwidth]{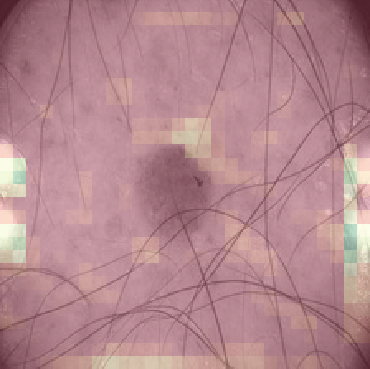} &
			\includegraphics[width = 0.21\textwidth]{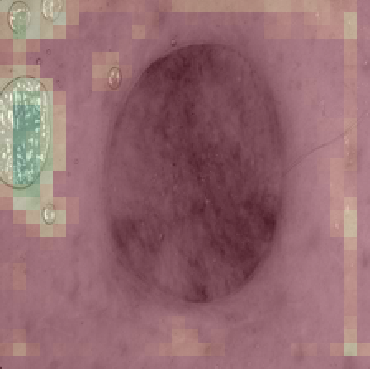} &
			\includegraphics[width = 0.21\textwidth]{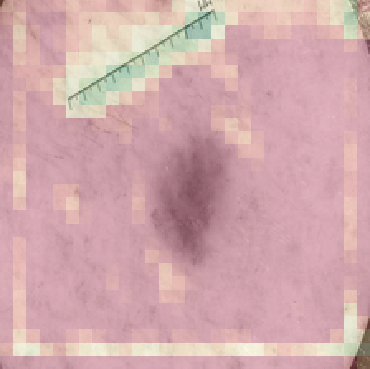} 
		\end{tabular}
		\caption{Feature maps with high activations on various image artifacts. Examples are, from left to right, specular reflection, gel treatment and rulers. These artifacts could potentially impose a bias on the output of the CNN. \label{fig:artifacts}}
	\end{center}
\end{figure}

\textbf{Artifacts} We also noticed that some of the feature maps have high activations on various artifacts in the images. For example, as shown in Fig.~\ref{fig:artifacts}, some feature maps have high activations on specular reflections, gel application, or rulers. This highlights some of the risks when using machine learning techniques, as this could impose a potential bias to the output of the network, when such artifacts are prominently present in the training images of a specific class.

\begin{figure}
\centering
\makebox[\textwidth][c]{\includegraphics[width=1.25\textwidth]{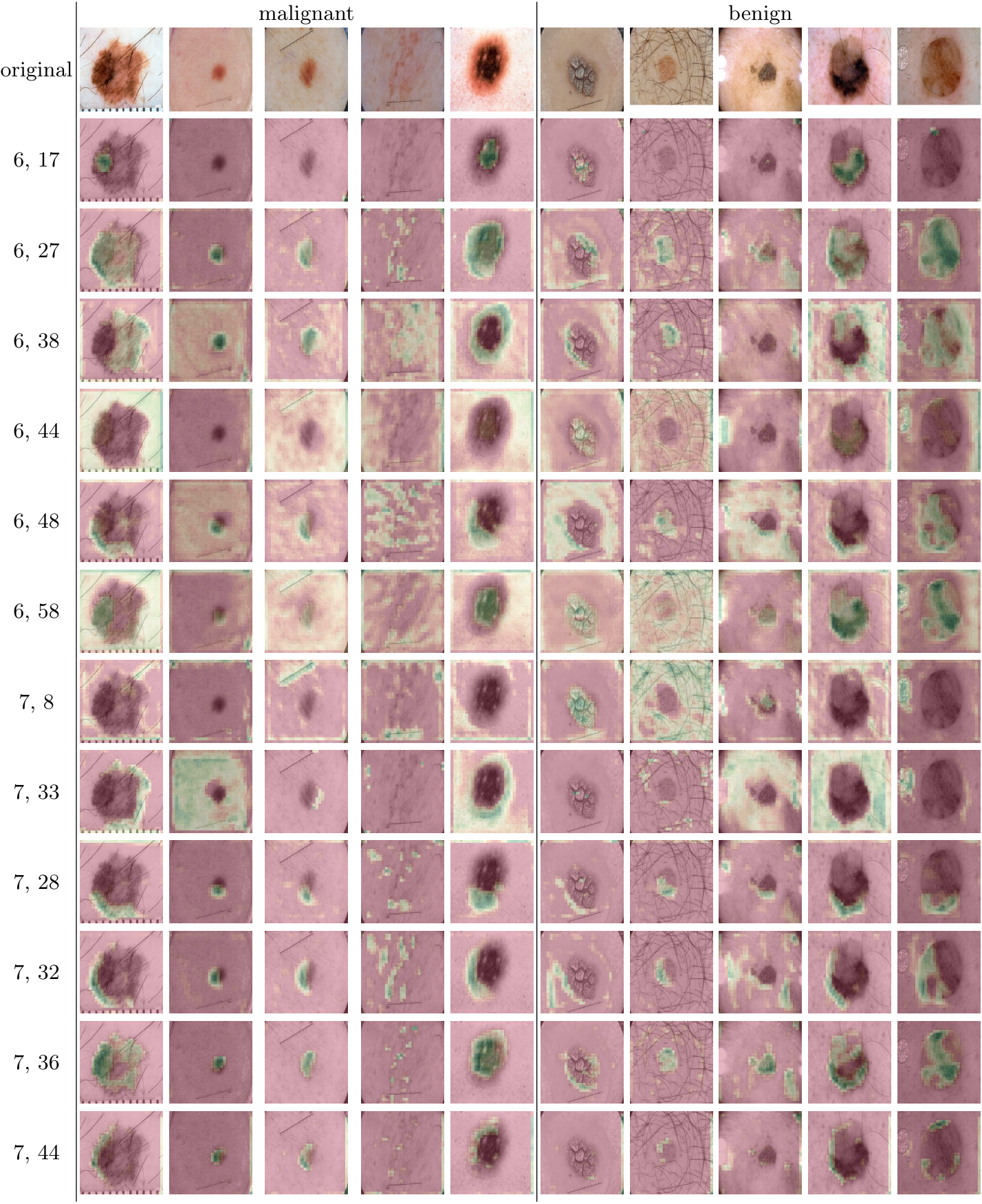}}
\caption{An overview of the feature maps. \label{fig:matrix}}
\end{figure}

A more elaborate overview of the activations of different feature maps on different images is shown in Fig.~\ref{fig:matrix}.

\section{Conclusion}

In this paper, we analyzed the features learned by a CNN, trained for skin lesion classification, in the field of dermatology. By visualizing the feature maps of the CNN, we see that, indeed, the high-level convolutional layers activate on similar concepts as used by dermatologists, such as lesion border, darker regions inside the lesion, surrounding skin, etc. We also found that some feature maps activate on various image artifacts, such as specular reflections, gel application, and rulers. This flags that one should be cautious when constructing a dataset for training, that such artifacts do not lead to a bias in the machine learning model.

Although this paper gives some insight in the features learned by the CNN, this does not yet explain any causal relation between the detected features of the CNN and its output. Furthermore, going through the feature maps, we did not find any that precisely highlight many of the other structures that dermatologists scan for, such as globules, dots, blood vessel structures, etc. We believe more research is required in this area in order to make CNNs a better decision support tool for dermatologists.

\clearpage

\bibliographystyle{splncs04}
\bibliography{bibtex}

\end{document}